\documentclass[10pt,twocolumn,letterpaper]{article}

\usepackage{template/iccv}
\usepackage{times}
\usepackage{epsfig}
\usepackage{amsmath}
\usepackage{amssymb}
\usepackage{graphicx}

\usepackage{mathtools}
\usepackage[titlenumbered,ruled,linesnumbered]{algorithm2e}
\usepackage{multirow}
\usepackage{algorithmicx}
\usepackage{subcaption}

\usepackage{lipsum}  
\usepackage{tikz}
\usepackage{pgfplots}
\usepackage{xspace}
\pgfplotsset{compat=1.9}
\usepgfplotslibrary{colorbrewer}
\pgfplotsset{cycle list/Dark2}

\usetikzlibrary{shapes.geometric,decorations.pathreplacing,calc}
\usepackage{xcolor}
\definecolor{gcn}{HTML}{9fbfff}
\definecolor{smooth}{HTML}{ffdf9f}
\definecolor{act}{HTML}{ffaf9f}
\definecolor{conv}{HTML}{FFA500}
\definecolor{pool}{HTML}{B22222}
\definecolor{up}{HTML}{B22222}
\definecolor{view}{HTML}{FFFFFF}
\definecolor{bn}{HTML}{FFD700}
\definecolor{brewgreen}{HTML}{66A61E}
\tikzset{smooth/.style={black,draw=black,fill=smooth,rectangle,minimum height=0.75cm}}
\tikzset{gcn/.style={black,draw=black,fill=gcn,rectangle,minimum height=1cm}}
\tikzset{act/.style={black,draw=black,fill=act,rectangle,minimum height=0.75cm}}
\tikzset{view/.style={black,draw=black,fill=view,rectangle,minimum height=0.75cm}}
\tikzset{up/.style={black,draw=black,fill=up,rectangle,minimum height=0.75cm}}
\tikzset{pool/.style={black,draw=black,fill=pool,rectangle,minimum height=0.75cm}}
\tikzset{bn/.style={black,draw=black,fill=bn,rectangle,minimum height=0.75cm}}
\tikzset{gbox/.style={black,draw=black,fill=green,rectangle,minimum height=0.75cm}}
\tikzset{rbox/.style={black,draw=black,fill=red,rectangle,minimum height=0.75cm}}
\tikzset{wbox/.style={black,draw=black,fill=white,rectangle,minimum height=0.75cm}}

\newcommand{\ccrl}{CCRL\xspace}

\newcommand{\trainSet}{$\mathcal{S}_\text{train}$\xspace}

\newcommand{\testSet}{$\mathcal{S}_\text{val}$\xspace}

\newcommand{\xij}{$\mathbf{x}_{i:j}^v$\xspace}

\usepackage[pagebackref=true,breaklinks=true,letterpaper=true,colorlinks,bookmarks=false]{hyperref}

\iccvfinalcopy 

\def\iccvPaperID{2} 

\ificcvfinal\fi
\begin{document}

\title{Cross-Class Relevance Learning for Temporal Concept Localization}

\author{
Junwei Ma~\thanks{Authors contributed equally to this work.}\\
Layer6 AI\\
{\tt\small jeremy@layer6.ai}
\and
Satya Krishna Gorti~\footnotemark[1]\\
Layer6 AI\\
{\tt\small satya@layer6.ai}
\and
Maksims Volkovs\\
Layer6 AI\\
{\tt\small maks@layer6.ai}
\and
Ilya Stanevich\\
Layer6 AI\\
{\tt\small ilya@layer6.ai}
\and
Guangwei Yu\\
Layer6 AI\\
{\tt\small guang@layer6.ai}
}

\maketitle

\begin{abstract}
We present a novel Cross-Class Relevance Learning approach for the task of
temporal concept localization. Most localization architectures rely on
feature extraction layers followed by a classification layer which outputs class probabilities
for each segment. However, in many real-world applications classes can exhibit complex 
relationships that are difficult to model with this architecture. In contrast, we propose 
to incorporate target class and class-related features as input, and learn a pairwise binary 
model to predict general segment to class relevance. This facilitates
learning of shared information between classes, and allows for arbitrary 
class-specific feature engineering. We apply this approach to the 3rd YouTube-8M Video 
Understanding Challenge together with other leading models, and achieve first 
place out of over 280 teams. In this paper we describe our approach and show
some empirical results.

\end{abstract}

\section{Introduction}\label{intro}

The problem of video understanding has received increasing 
attention recently following the rapid advancement in 
computer vision image models. Temporal concept localization in
particular is a challenging but fundamental problem in 
video understanding where the main goal is to identify video 
segments that contain a specific concept or action.
Just as the rapid advancement in image understanding is
aided by the large-scale and diverse
ImageNet dataset~\cite{deng2009imagenet}, the recently
introduced YouTube-8M dataset~\cite{abu2016youtube}
brings both scale and diversity to video data. The 3rd 
YouTube-8M Video Understanding 
Challenge~\footnote{\url{www.kaggle.com/c/youtube8m-2019}}
leverages this data to benchmark video understanding
models in a standardized setting.

Video action localization is a closely related problem to concept localization. 
Action localization generally assumes that action classes are mutually exclusive, 
meaning that each segment can only contain one class. However, in the real world applications 
with many classes, a segment may contain multiple actions simultaneously.
Moreover, in addition to actions there are also objects of interest that need to be
accurately localized. This complexity is captured by the diverse set of classes in the 
YouTube-8M dataset that include both actions and objects, and where segments
are labelled with multiple classes introducing new modelling challenges. At scale, 
the total number of segment-class pairs becomes prohibitively large. So in this data
only a small subset is sampled and labelled. The labels are binary where 1 indicates that
segment contains a given class, and 0 indicates that it doesn't contain it. The goal
of the challenge is to build accurate segment models that, given a video, can identify
all segments that contain each of the classes.

Existing approaches primarily apply recurrent~\cite{miech2017learnable} or
codebook-based~\cite{skalic2018building} models to extract video and segment features. 
Then a fully connected classification layer predicts class probabilities for
each segment. The idea is to use a common feature extraction backbone, and learn class-specific 
weights in the last layer. This architecture makes it challenging to incorporate 
any class information such as hierarchy or class similarity into the model. 
Furthermore, when number of classes is large and/or few training examples are available per
class, this model is prone to over-fitting as it learns a separate set of weights for every class. To deal with these problems we propose a pairwise segment-class model that in addition to segment also takes as input target class and class-related features, and predicts segment to class relevance. This facilitates learning of shared information between classes, and allows 
for arbitrary class-specific feature engineering. 

To avoid running expensive inference for every segment-class pair we further 
propose a candidate generation pipeline. Specifically, we apply video level model
to significantly reduce number of candidate segments for each class while 
preserving high recall. To summarize, our contribution in this paper is two-fold. 
First, we propose a novel architecture for video concept localization that 
incorporates target class information as input into the model. We further develop 
class-specific features that improve localization performance. Second, we demonstrate  
practical and efficient application of the proposed architecture on the largest video
dataset available to date as part of the 3rd YouTube-8M Video Understanding 
Challenge. Our approach achieves first place out of over 280 teams.

\section{Related Work}

	The temporal concept localization task is closely related to video action classification and 
	action localization. Video action classification task consists of predicting what actions 
	appear in a given video, while action localization aims to find segments where each action appears. 
	However, classes in the YouTube-8M dataset are more diverse than actions, and include everyday
	objects that exhibit hierarchy and relationships not found in traditional action
	classification and localization datsets~\cite{caba2015activitynet,idrees2017thumos,yeung2018every}.
	
	\textbf{Feature extraction}. Feature extraction is a common step in both tasks. Early works extract
	hand-crafted features by tracking pixels over time~\cite{wang2013dense}. Recent works instead 
	employ deep neural networks that use spatio-temporal convolutions such as I3D \cite{carreira2017quo},
	C3D \cite{Tran_2015_ICCV} and two-stream approaches \cite{feichtenhofer2016convolutional,carreira2017quo}
	to capture visual and motion features. The YouTube-8M video and segment 
	datasets~\cite{abu2016youtube} used in our experiments
	consist of both video and audio features extracted from publicly available Inception network~\cite{szegedy2016rethinking} and VGG-inspired acoustic model~\cite{45611} respectively.
	
	\textbf{Action Classification}. 
	Recurrent models such as LSTMs \cite{gers1999learning} and GRUs \cite{cho2014learning}
	are natural candidates for video level action classification as they are well suited to extract temporal features across time. Applications of 
	recurrent models on this task can be seen in~\cite{skalic2018building} and~\cite{DBLP:journals/corr/abs-1810-00207}.
	Other popular approaches for action classification~\cite{Kantorov_2014_CVPR,peng2014boosting} are based on Fisher
	Vectors (FV) \cite{perronnin2007fisher} and Vector of Locally aggregated Descriptors (VLAD)~\cite{jegou2010aggregating}. These 
	methods rely on maintaining a hand-crafted codebook. NetVLAD~\cite{arandjelovic2016netvlad} and NetFV~\cite{miech2017learnable} 
	are variations of VLAD and FV methods that learn the encoding end-to-end using deep networks. These methods achieve 
	state-of-the-art performance on the YouTube-8M video classification task~\cite{miech2017learnable,skalic2018building,DBLP:journals/corr/abs-1810-00207,DBLP:journals/corr/abs-1809-04403}, and we also leverage them in our pipeline.
	
	\textbf{Action Localization}. Action localization task can be partitioned into two groups: \textit{fully-supervised} and \textit{weakly-supervised}. In the fully supervised setting, models are trained using frame-level annotations indicating action
	boundaries. \cite{chao2018rethinking,shou2017cdc} perform fully supervised action localization is a two stage manner, first predicting proposals and
	further classifying proposals into action classes. GTAN~\cite{long2019gaussian} learns a set of Gaussian kernels used to predict intervals of actions in a
	single-stage manner. The network is optimized end-to-end using classification loss and two regression losses to predict action boundaries.
	
	On the other hand, weakly supervised action localization models predict action classes and boundaries only based on high level action category labels. This
	can be formulated as multi-instance multi-label learning problem~\cite{zhou2012multi}, where multiple instances describe an example having multiple
	labels. For example, W-TALC \cite{paul2018w} extends the work of STPN \cite{nguyen2018weakly} which introduces attention module to identify sparse subset of
	video frames associated with actions by having novel objective functions based on multiple-instance learning and activity similarity.
	
	\textbf{Datasets}. Some notable datasets for action classification include Moments in Time \cite{monfort2019moments}, Something-Something
	\cite{goyal2017something}, UCF101 \cite{soomro2012dataset}, Sports1M \cite{karpathy2014large}. Thumos \cite{idrees2017thumos}, MultiThumos
	\cite{yeung2018every} and ActivityNet \cite{caba2015activitynet} are popular datasets used in recent work for action localization. 
	Another category of datasets used for spatio-temporal localization include MSRActions \cite{yuan2009discriminative}, 
	AVA \cite{gu2018ava} and UCFSports \cite{rodriguez2008action}. Compared to these, Youtube-8M \cite{abu2016youtube} is by
	far the largest multi-class dataset with over 3.8K classes that contain a very broad 
	and diverse set of actions and objects.
	
	\textbf{Pairwise Relevance Learning}. 
	It is common in structured prediction tasks to design a pairwise compatibility function over input-output pairs~\cite{tsochantaridis2005large}. This can be seen for example in \cite{Akata_2013_CVPR}, where joint image-class model is trained to embed classes 
	and measure compatibility between images and classes for zero-shot 
	classification. Our approach follows similar intuition, and we aim to learn
	a general pairwise model to predict segment to class relevance.

\section{Approach}

Given a video $v$, we use $\mathbf{x}_i^v$ to denote 
the $i$'th frame in $v$. We follow~\cite{abu2016youtube}
and represent video by a sequence of frames 
$\mathbf{x}^v=[\mathbf{x}_1^v,\mathbf{x}_2^v,...,
\mathbf{x}_{F_v}^v]$, where $F_v$ is the total number of frames in $v$. 
We assume that features have been extracted for each frame so 
$\mathbf{x}_i^v$ is a vector in $\mathcal{R}^d$.
Segment $\mathbf{x}_{i:j}^v=[\mathbf{x}_i^v,\mathbf{x}_{i+1}^v,...,
\mathbf{x}_{j}^v]$ is a sub-sequence of consecutive frames from
$i$ to $j$ with $j > i$.
The temporal concept localization problem is formulated as
predicting relevance of a given segment \xij to class $c$. 
The ground truth binary label for this 
task is denoted by $y_{v,i:j}^c\in\{0,1\}$, where
positive label $1$ indicates that \xij is relevant
to $c$ and negative label $0$ indicates that \xij is
not relevant to $c$. For a large database of videos and classes, 
it's impractical to label all segment-class pairs.
Typically only a subset of pairs is sampled and labelled, so 
$y_{v,i:j}^c$ is unknown for most segments \xij.
In the following sections we outline our approach 
to this problem and show empirical results.
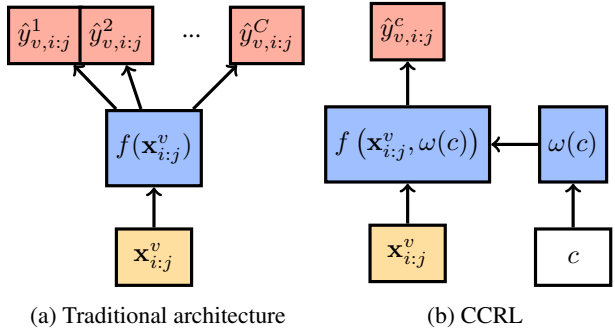
\begin{figure}[t]
    \begin{subfigure}[b]{0.23\textwidth}
        \scalebox{1}{%
        \begin{tikzpicture}[line width=1.2pt]
            \node[smooth,minimum width=1cm] (seq0) at (0.0,1.55) {$\mathbf{x}^v_{i:j}$};
            \node[gcn] (f) at (0,3) {$f(\mathbf{x}^v_{i:j})$};
            \node[act] (y0) at (-1.45,4.5) {$\hat{y}_{v,i:j}^1$};
            \node[act] (y1) at (-0.5,4.5) {$\hat{y}_{v,i:j}^2$};
            \node[] (ydot) at (0.5,4.5) {...};
            \node[act] (yc) at (1.5,4.5) {$\hat{y}_{v,i:j}^C$};
            \draw[->] (seq0) -- (f);
            \draw[->] (f) -- (y0);
            \draw[->] (f) -- (y1);
            \draw[->] (f) -- (yc);
        \end{tikzpicture}
        }
        \caption{Traditional architecture}
        \label{fig:classifier}
     \end{subfigure}%
     \hskip 0.2cm
    \begin{subfigure}[b]{0.23\textwidth}
        \scalebox{1}{%
        \begin{tikzpicture}[line width=1.2pt]
            \node[view,minimum width=1cm] (c) at (2.2,1.5) {$c$};
            \node[smooth,minimum width=1cm] (seq0) at (0.0,1.55) {$\mathbf{x}^v_{i:j}$};
            \node[gcn] (f) at (0,3) {$f\left(\mathbf{x}_{i:j}^v, \omega(c)\right)$};
            \node[gcn] (omega) at (2.2,3) {$\omega(c)$};
            \node[act] (yc) at (0,4.5) {$\hat{y}_{v,i:j}^c$};
            \draw[->] (c) -- (omega);
            \draw[->] (seq0) -- (f);
            \draw[->] (f) -- (yc);
            \draw[->] (omega) -- (f);
        \end{tikzpicture}
        }
        \caption{CCRL}
        \label{fig:joint_rel}
     \end{subfigure}
  \caption{(\subref{fig:classifier}) shows commonly used classification
  architecture where a vector of class probabilities is predicted for 
  each segment. (\subref{fig:joint_rel}) shows our proposed
  CCRL architecture where segment \xij and class features $\omega(c)$
  are jointly used as input to a binary relevance model.
  }\label{fig:arch}
  \vskip -0.5cm
\end{figure}

\subsection{Candidate Generation}\label{sec:candidate_gen}

At scale, the number of segments can become prohibitively large
making model inference very expensive. To address this, we propose a 
candidate generation stage to significantly reduce the set of segments 
that need to be scored for each class. We use video level 
candidate generation and select videos that 
are likely to contain segments with a given class. The main motivation 
for this is that predicting whether video contains a class is 
significantly easier than localizing where exactly it appears.
This is evidenced by the performance of video models from 
the previous edition of the YouTube-8M challenge that focused on
video classification. Winning solutions were able to achieve nearly
0.9 in mean average precision on this 
task~\cite{skalic2018building} even though there were over 3.8K classes.
Moreover, labelling videos is significantly easier than labelling 
segments with precise start and end times. YouTube-8M dataset has
video level labels for more than 6M videos but only 
47K videos have segment labels. Using video models thus also allows us to 
leverage information from the much larger video dataset and transfer
it to segment prediction.

Our candidate generation stage uses a video level model to 
predict class probabilities for every video in the corpus.
Then, given a target class $c$, we sort videos according to
probability for $c$ and take top-K. All segments from the top-K
videos are treated as candidates for $c$ and are passed to the segment 
model. By leveraging leading approaches for video 
classification~\cite{skalic2018building} we are able to reduce
the number of candidate segments by over 20x while still achieving 
near perfect $99.7\%$ average segment recall across classes. 
\begin{figure}[t]\centering
    \input{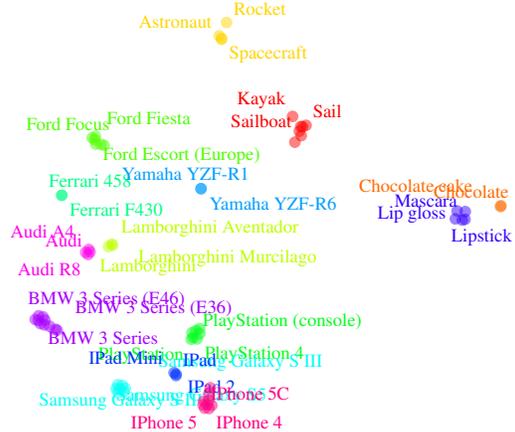}
  \caption{t-SNE reduced bag-of-words description vectors 
  for a subset of Youtube-8M classes. Different colors represent 
  clusters found by applying k-means. Clear 
  hierarchical relationships between classes can
  be observed from this diagram.
  }\label{fig:hierarchy}
\end{figure}

\subsection{Traditional Approach to Classification}\label{sec:baseline}

Typical architecture for segment/video classification consists of feature 
extraction backbone followed by a classification layer where a separate set of
weights is learned for each class. The main idea is that the backbone
learns to extract visual information generally useful for prediction, 
and classification layer learns to transform this information into 
class-specific predictions. Formally, given a segment \xij the model 
outputs a prediction vector:
\begin{equation}\label{eq:baseline}
    {\bf \hat{y}}_{v,i:j} = f(\mathbf{x}^v_{i:j})
\end{equation}
where ${\bf \hat{y}}_{v,i:j} = 
[\hat{y}^1_{v,i:j}, \hat{y}^2_{v,i:j},...,\hat{y}^C_{v,i:j}]$ contains
predictions for all target classes. This architecture is illustrated in
Figure~\ref{fig:classifier}.

Despite its popularity, this architecture has several disadvantages.
First, in many real-world applications classes often have close 
hierarchical relationships. An example of this  
is shown in Figure~\ref{fig:hierarchy}. Here, we show
t-SNE~\cite{maaten2008visualizing} visualization of the bag-of-words 
descriptions for a subset of classes in YouTube-8M. From the figure
we can observe complex and hierarchical inter-class relationships 
common in these domains. For example, a number of classes are related to 
automobiles. These are sub-divided into economy, sports and luxury, 
and can be further partitioned by manufacturer etc. Incorporating this information or 
any other class related features is non-trivial in the traditional model. 
It typically requires loss modification and/or manual weight manipulation
in the classification layer~\cite{Akata_2013_CVPR}. Second, class-specific 
weights are only updated when examples of that class are encountered 
during training. For classes with few training examples this can result 
in under/over-fitted models. Accurately labelling video segments is a difficult task
and most datasets including YouTube-8M have highly sparse labels.

In the following section we propose a different architecture for 
segment classification that avoids most of these problems.
\begin{table}[t]
\caption{Performance of the segment candidate generation models evaluated
by Recall at 100K segments for each class.}
\centering
\begin{tabular}{ |p{3cm}| p{2.5cm}| }
 \hline
 Model      &   Recall@$100K$ \\
 \hline
 Logistic  &   0.9713 \\
 LSTM            &   0.9931  \\
 Ensemble~\cite{skalic2018building}   &   \textbf{0.9969}   \\
 \hline
\end{tabular}
\label{table:candidate_recall}
\end{table}

\subsection{Cross-Class Relevance Learning}\label{sec:ccrl}

To incorporate class information into the model and promote 
cross-class information sharing, we propose a pairwise architecture.
Specifically, our architecture takes both segment and target class
as input and predicts relevance probability. We train a {\it single} 
pairwise model that captures general notion of ``relevance" between 
all segments and classes. This architecture is shown in Figure~\ref{fig:joint_rel}.
Using $\omega(c)$ to denote extracted features for class $c$, our
model outputs relevance probability between segment \xij and $c$:
\begin{equation}\label{eq:ccrl_classifier}
    \hat{y}_{v,i:j}^c = f\left(\mathbf{x}^v_{i:j}, \omega(c)\right)
\end{equation}
Pairwise architecture allows us to encode arbitrary class information
into the model in a principled way. By training a single
model for all classes we can automatically learn relationships between
classes that are useful for the target task. To further illustrate this
point, consider an example where $f$ is a decision tree model and $\omega(c)$ simply 
outputs class index $c$. The decision tree can use class index to group
classes in each sub-tree and automatically learn
an implicit hierarchy. This is not possible with traditional 
classification architectures. Another advantage is that a learned
model can potentially generalize to new unseen classes during training 
as long as $\omega(c)$ can reasonably encode them.

We train this model with a binary cross entropy objective using 
all labelled segments:
\begin{equation}
\begin{aligned}
\mathcal{L} = &-\sum_{y_{v,i:j}^c} 
y_{v,i:j}^c \log \left(f(\mathbf{x}^v_{i:j}, \omega(c))\right)\\
&+(1 - y_{v,i:j}^c) \log \left(1 - f(\mathbf{x}^v_{i:j}, \omega(c))\right)
\end{aligned}
\end{equation}
During inference, given a segment \xij we need to determine which classes 
are relevant for \xij. Naive approach requires $C$ forward passes through the model,
one for every class. This quickly becomes prohibitively expensive, especially for 
datasets such as YouTube-8M that have thousands of classes. However, we
note that after candidate generation step we only need to consider a small
subset of segments for each class. The two-stage approach makes inference 
in our model practical and scalable, and we run it without difficulties on
the large scale YouTube-8M data.
\begin{table}[t]
\centering
\caption{Private leaderbooard performance of segment models with and without
candidate generation (CG)}
\begin{tabular}{ |l|c|c|}
\hline
 Model      & Without CG &  With CG \\
 \hline
 ConvNet \cite{bai2018empirical} & 0.7454 &     0.8036 \\
 LSTM \cite{skalic2018building} & 0.7681                  & 0.8023 \\
 Transformer \cite{Vaswani2017AttentionIA} & 0.6524    & 0.7955 \\
 NetVLAD \cite{arandjelovic2016netvlad} & 0.7243         & 0.8023 \\
 NetFV \cite{miech2017learnable} & 0.7203      & 0.8028 \\
 \ccrl& \textbf{0.7711}  & \textbf{0.8091}  \\
 \hline
 Ensemble  & - & {\bf 0.8329}\\
\hline
\end{tabular}
\label{table:segment_rerank}
\end{table}

\subsection{Class Features}\label{sec:features}

We conducted extensive investigation on the types of class features
that can be encoded with $\omega(c)$. Here, we present one 
interesting feature that significantly improves performance.
Given a segment \xij and target class $c$, our goal is to compare 
\xij to all labelled relevant and irrelevant segments for class $c$.
The main idea is similar to clustering in that if \xij is relevant for  
$c$ then it should be ``close" to other relevant segments and ``far"
from irrelevant ones. We formalize this as follows:
\begin{align*}\label{eq:sim}
    \text{SIM}_{\text{pos}}(\mathbf{x}_{i:j}^v,c) &= \sum_{u\neq v}\sum_{y_{u,k:l}^c=1}
    \text{dist}(\mathbf{x}^v_{i:j}, \mathbf{x}^u_{k:l})\\
    \text{SIM}_{\text{neg}}(\mathbf{x}_{i:j}^v,c) &= \sum_{u\neq v}\sum_{y_{u,k:l}^c=0}
    \text{dist}(\mathbf{x}^v_{i:j}, \mathbf{x}^u_{k:l})
\end{align*}
where  $\text{dist}(\mathbf{x}^v_{i:j}, \mathbf{x}^u_{k:l})$ is a distance
function between segments \xij and $\mathbf{x}^u_{k:l}$.
$\text{SIM}_{\text{pos}}$ and $\text{SIM}_{\text{neg}}$ compute 
total distances to relevant (positive) and irrelevant (negative) segments
respectively. Any similarity function can be used for dist. Empirically, we
found cosine of the angle between segment feature encodings 
(e.g. RNN hidden states) to work well. Adding these features 
significantly improved localization accuracy, and further demonstrates
the flexibility of the proposed architecture that allows to explore
virtually any class-related features.

\section{Experiments}\label{sec:experiments}

We conduct experiments on temporal localization task
as part of the 3rd YouTube-8M Challenge. This challenge 
dataset is an extension of the original 
YouTube-8M dataset~\cite{abu2016youtube}, which contains
6.1M videos described by 2.6B audio-visual frame
features. The videos are labelled across 3862 classes,
and have an average of 3 labels per video. The extended
dataset contains 237K human-verified segment class labels
from 47K videos with approximately 5 segments per video. The videos
in the extended dataset are labelled with 1000 classes
which is a subset of the original 3862 classes.
\begin{table}[t]
\centering
\caption{Private leaderboard performance for top-5 teams 
as well as our individual models. Our team ``Layer6 AI"
achieved first place.
}
\begin{tabular}{ |l|c|c|}
 \hline
 Team      & Leaderboard &  Rank \\
 \hline
 Layer6 AI   &   \textbf{0.8329}  & \textbf{1} \\
 BigVid Lab  &   0.8262 & 2 \\
 RLin  &   0.8255         & 3\\
 bestfitting &      0.8171 & 4 \\
 Last Top GB Model &   0.8046    & 5 \\
 \hline
 ConvNet & 0.8036 & 6 \\
 LSTM            &   0.8023         & 7 \\
 Transformer & 0.7955 & 8 \\
 NetVLAD & 0.8023 & 7 \\
 NetFV & 0.8028 & 7 \\
 \ccrl & {\bf 0.8091} & {\bf 5} \\
\hline
\end{tabular}
\label{table:ranks}
\end{table}

Challenge data is split into a publicly available labelled set and two
held-out test sets (public and private leaderboards). For all experiments, 
we split the labelled set into $90\%$ \trainSet and $10\%$ \testSet 
sets and tune hyperparameters using \testSet set. We use the chosen
hyperparameters to the re-train on the full $100\%$ labelled set. 
Results for the held-out set are reported by submitting our model 
predictions to the leaderboard through the Kaggle platform. We 
report results on the private leaderboard which corresponds to 
the larger held-out test set that was used to rank teams at the
end of the competition.

The challenge task is to predict a ranked list of up to 100K segments for 
each class. This list is evaluated against the ground truth using
mean average precision (mAP):
\begin{equation}
    \text{mAP} = \frac{1}{C} \sum^{C}_{c=1}\frac{\sum^{n}_{i=1}{\text{Prec}(i) \times rel(i)}}{N_c}
\end{equation}
where $\text{Prec}(i)$ is precision at rank $i$ and
$rel(i)$ is 1 if segment at rank $i$ is relevant
and 0 otherwise; $N_c$ denotes the total number of relevant segments
in class $c$.
In addition to mAP, we use average class recall to evaluate the quality 
of the segment candidate generation model.

\subsection{Implementation}
We compare popular temporal localisation models against our proposed \ccrl
architecture. To make comparison fair all models share the same
candidate segment generation pipeline.
\begin{figure}[t]\centering
\begin{tikzpicture}

\definecolor{color1}{rgb}{1,0.498039215686275,0.0549019607843137}
\definecolor{color0}{rgb}{0.12156862745098,0.466666666666667,0.705882352941177}

\begin{axis}[
height=5cm,
width=8cm,
legend cell align={left},
legend style={draw=white!80.0!black},
tick align=outside,
tick pos=both,
x grid style={white!69.01960784313725!black},
xmin=-1.8, xmax=26.8,
xtick style={color=black},
xtick={0,1,2,3,4,5,6,7,8,9,10,11,12,13,14,15,16,17,18,19,20,21,22,23,24,25},
xticklabel style = {rotate=90.0},
xticklabels={BMW M3,BMW 3 Series (E30),BMW 3 Series (E36),BMW 3 Series (E46),,Samsung Galaxy S5,Samsung Galaxy S4,Samsung Galaxy,Samsung Galaxy Note II,,Cell (Dragon Ball),Gohan,Majin Boo,Freeza,,Boeing 737,Boeing 777,Airbus A380,Boeing 747,,Xbox,Xbox (console),Xbox 360,Xbox One,Kinect,},
y grid style={white!69.01960784313725!black},
ymin=0, ymax=1.025370370395,
ytick style={color=black},
ytick={0,0.2,0.4,0.6,0.8,1,1.2},
yticklabels={0.0,0.2,0.4,0.6,0.8,1.0,1.2},
xtick pos=left,
ytick pos=left,
ylabel=AP,
]
\draw[fill=color0,draw opacity=0,fill opacity=0.8] (axis cs:-0.5,0) rectangle (axis cs:0.5,0.9715909091);
\addlegendimage{ybar,ybar legend,fill=color0,draw opacity=0,fill opacity=0.8};
\addlegendentry{CCRL}

\draw[fill=color0,draw opacity=0,fill opacity=0.8] (axis cs:0.5,0) rectangle (axis cs:1.5,0.9666666667);
\draw[fill=color0,draw opacity=0,fill opacity=0.8] (axis cs:1.5,0) rectangle (axis cs:2.5,0.9620159753);
\draw[fill=color0,draw opacity=0,fill opacity=0.8] (axis cs:2.5,0) rectangle (axis cs:3.5,0.9765432099);
\draw[fill=color0,draw opacity=0,fill opacity=0.8] (axis cs:3.5,0) rectangle (axis cs:4.5,0);
\draw[fill=color0,draw opacity=0,fill opacity=0.8] (axis cs:4.5,0) rectangle (axis cs:5.5,0.9159497069);
\draw[fill=color0,draw opacity=0,fill opacity=0.8] (axis cs:5.5,0) rectangle (axis cs:6.5,0.9261387327);
\draw[fill=color0,draw opacity=0,fill opacity=0.8] (axis cs:6.5,0) rectangle (axis cs:7.5,0.9666666667);
\draw[fill=color0,draw opacity=0,fill opacity=0.8] (axis cs:7.5,0) rectangle (axis cs:8.5,0.8615015284);
\draw[fill=color0,draw opacity=0,fill opacity=0.8] (axis cs:8.5,0) rectangle (axis cs:9.5,0);
\draw[fill=color0,draw opacity=0,fill opacity=0.8] (axis cs:9.5,0) rectangle (axis cs:10.5,0.8951564517);
\draw[fill=color0,draw opacity=0,fill opacity=0.8] (axis cs:10.5,0) rectangle (axis cs:11.5,0.8590474173);
\draw[fill=color0,draw opacity=0,fill opacity=0.8] (axis cs:11.5,0) rectangle (axis cs:12.5,0.8574945165);
\draw[fill=color0,draw opacity=0,fill opacity=0.8] (axis cs:12.5,0) rectangle (axis cs:13.5,0.8165523848);
\draw[fill=color0,draw opacity=0,fill opacity=0.8] (axis cs:13.5,0) rectangle (axis cs:14.5,0);
\draw[fill=color0,draw opacity=0,fill opacity=0.8] (axis cs:14.5,0) rectangle (axis cs:15.5,0.8036699675);
\draw[fill=color0,draw opacity=0,fill opacity=0.8] (axis cs:15.5,0) rectangle (axis cs:16.5,0.5654761905);
\draw[fill=color0,draw opacity=0,fill opacity=0.8] (axis cs:16.5,0) rectangle (axis cs:17.5,0.828710967);
\draw[fill=color0,draw opacity=0,fill opacity=0.8] (axis cs:17.5,0) rectangle (axis cs:18.5,0.8014818515);
\draw[fill=color0,draw opacity=0,fill opacity=0.8] (axis cs:18.5,0) rectangle (axis cs:19.5,0);
\draw[fill=color0,draw opacity=0,fill opacity=0.8] (axis cs:19.5,0) rectangle (axis cs:20.5,0.5);
\draw[fill=color0,draw opacity=0,fill opacity=0.8] (axis cs:20.5,0) rectangle (axis cs:21.5,0.8714285714);
\draw[fill=color0,draw opacity=0,fill opacity=0.8] (axis cs:21.5,0) rectangle (axis cs:22.5,0.5);
\draw[fill=color0,draw opacity=0,fill opacity=0.8] (axis cs:22.5,0) rectangle (axis cs:23.5,0.6428571429);
\draw[fill=color0,draw opacity=0,fill opacity=0.8] (axis cs:23.5,0) rectangle (axis cs:24.5,0.1666666667);
\draw[fill=color0,draw opacity=0,fill opacity=0.8] (axis cs:24.5,0) rectangle (axis cs:25.5,0);
\draw[fill=color1,draw opacity=0,fill opacity=0.8] (axis cs:-0.4,0) rectangle (axis cs:0.4,0.944077135);
\addlegendimage{ybar,ybar legend,fill=color1,draw opacity=0,fill opacity=0.8};
\addlegendentry{LSTM}

\draw[fill=color1,draw opacity=0,fill opacity=0.8] (axis cs:0.6,0) rectangle (axis cs:1.4,0.7683333333);
\draw[fill=color1,draw opacity=0,fill opacity=0.8] (axis cs:1.6,0) rectangle (axis cs:2.4,0.8808374776);
\draw[fill=color1,draw opacity=0,fill opacity=0.8] (axis cs:2.6,0) rectangle (axis cs:3.4,0.962654321);
\draw[fill=color1,draw opacity=0,fill opacity=0.8] (axis cs:3.6,0) rectangle (axis cs:4.4,0);
\draw[fill=color1,draw opacity=0,fill opacity=0.8] (axis cs:4.6,0) rectangle (axis cs:5.4,0.9072677323);
\draw[fill=color1,draw opacity=0,fill opacity=0.8] (axis cs:5.6,0) rectangle (axis cs:6.4,0.8870251651);
\draw[fill=color1,draw opacity=0,fill opacity=0.8] (axis cs:6.6,0) rectangle (axis cs:7.4,0.6592857143);
\draw[fill=color1,draw opacity=0,fill opacity=0.8] (axis cs:7.6,0) rectangle (axis cs:8.4,0.8414654661);
\draw[fill=color1,draw opacity=0,fill opacity=0.8] (axis cs:8.6,0) rectangle (axis cs:9.4,0);
\draw[fill=color1,draw opacity=0,fill opacity=0.8] (axis cs:9.6,0) rectangle (axis cs:10.4,0.6830879438);
\draw[fill=color1,draw opacity=0,fill opacity=0.8] (axis cs:10.6,0) rectangle (axis cs:11.4,0.8036036492);
\draw[fill=color1,draw opacity=0,fill opacity=0.8] (axis cs:11.6,0) rectangle (axis cs:12.4,0.766845014);
\draw[fill=color1,draw opacity=0,fill opacity=0.8] (axis cs:12.6,0) rectangle (axis cs:13.4,0.8085221325);
\draw[fill=color1,draw opacity=0,fill opacity=0.8] (axis cs:13.6,0) rectangle (axis cs:14.4,0);
\draw[fill=color1,draw opacity=0,fill opacity=0.8] (axis cs:14.6,0) rectangle (axis cs:15.4,0.7818810799);
\draw[fill=color1,draw opacity=0,fill opacity=0.8] (axis cs:15.6,0) rectangle (axis cs:16.4,0.4404761905);
\draw[fill=color1,draw opacity=0,fill opacity=0.8] (axis cs:16.6,0) rectangle (axis cs:17.4,0.6825088485);
\draw[fill=color1,draw opacity=0,fill opacity=0.8] (axis cs:17.6,0) rectangle (axis cs:18.4,0.7021320346);
\draw[fill=color1,draw opacity=0,fill opacity=0.8] (axis cs:18.6,0) rectangle (axis cs:19.4,0);
\draw[fill=color1,draw opacity=0,fill opacity=0.8] (axis cs:19.6,0) rectangle (axis cs:20.4,0.125);
\draw[fill=color1,draw opacity=0,fill opacity=0.8] (axis cs:20.6,0) rectangle (axis cs:21.4,0.6202564103);
\draw[fill=color1,draw opacity=0,fill opacity=0.8] (axis cs:21.6,0) rectangle (axis cs:22.4,0.3333333333);
\draw[fill=color1,draw opacity=0,fill opacity=0.8] (axis cs:22.6,0) rectangle (axis cs:23.4,0.5277777778);
\draw[fill=color1,draw opacity=0,fill opacity=0.8] (axis cs:23.6,0) rectangle (axis cs:24.4,0.05263157895);
\draw[fill=color1,draw opacity=0,fill opacity=0.8] (axis cs:24.6,0) rectangle (axis cs:25.4,0);
\end{axis}

\end{tikzpicture}
  \caption{Average precision performance for individual classes for CCRL and LSTM models.
  }\label{fig:ap}
  \vskip -0.5cm
\end{figure}
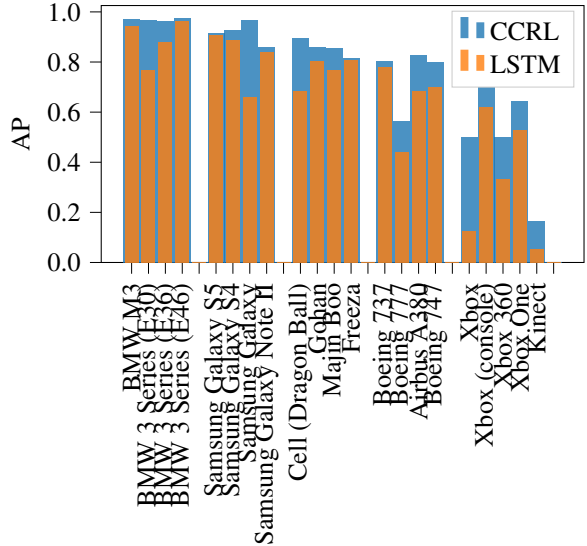

{\bf Candidate Generation Models} For segment candidate generation
we consider logistic and LSTM models that were part of the winning 
solution for the YouTube-8M video classification challenge~\cite{skalic2018building}.
Logistic model described in~\cite{abu2016youtube} is a 
fully-connected architecture with a sigmoid activation to output class 
probability predictions. The input features are composed 
by concatenating RGB and audio inputs for each frame.
LSTM model is a stack of bi-directional LSTM layers followed
by a uni-directional LSTM layer with a cell size of $1024$. A fully-connected
layer is then used for classification from the final hidden state 
of the uni-directional LSTM layer. We also evaluate model ensemble obtained
through knowledge distillation~\cite{skalic2018building}.

{\bf Segment Models}
For segment classification we consider latest neural network architectures
for temporal learning. These include Convolutional Neural Networks (ConvNet)~\cite{bai2018empirical}, LSTM~\cite{skalic2018building}, 
Transformer~\cite{Vaswani2017AttentionIA}, NetVLAD~\cite{arandjelovic2016netvlad} 
and NetFV~\cite{miech2017learnable}. In ConvNet model, two layers of width 2 convolutions
are applied along the video frames. Frame representations from last layer are 
then combined via average or max pooling and passed to classification layer.
Transformer model first compresses frames into segment representations using 
a convolutional layer, then three blocks of self-attention are applied followed 
by a fully connected classification layer. LSTM is initialised with pre-trained 
video LSTM model used for candidate generation. We then fine-tune it for localisation
task using segment level binary cross entropy loss.
NetVLAD~\cite{arandjelovic2016netvlad} and NetFV models are based on~\cite{miech2017learnable}.
We modify the original implementation by removing batch norm layers and switching from 
mixture-of-experts~\cite{abu2016youtube} to logistic regression. These modifications 
lead to more stable learning and better accuracy.
\begin{figure*}[t]\centering
\includegraphics[scale=0.4]{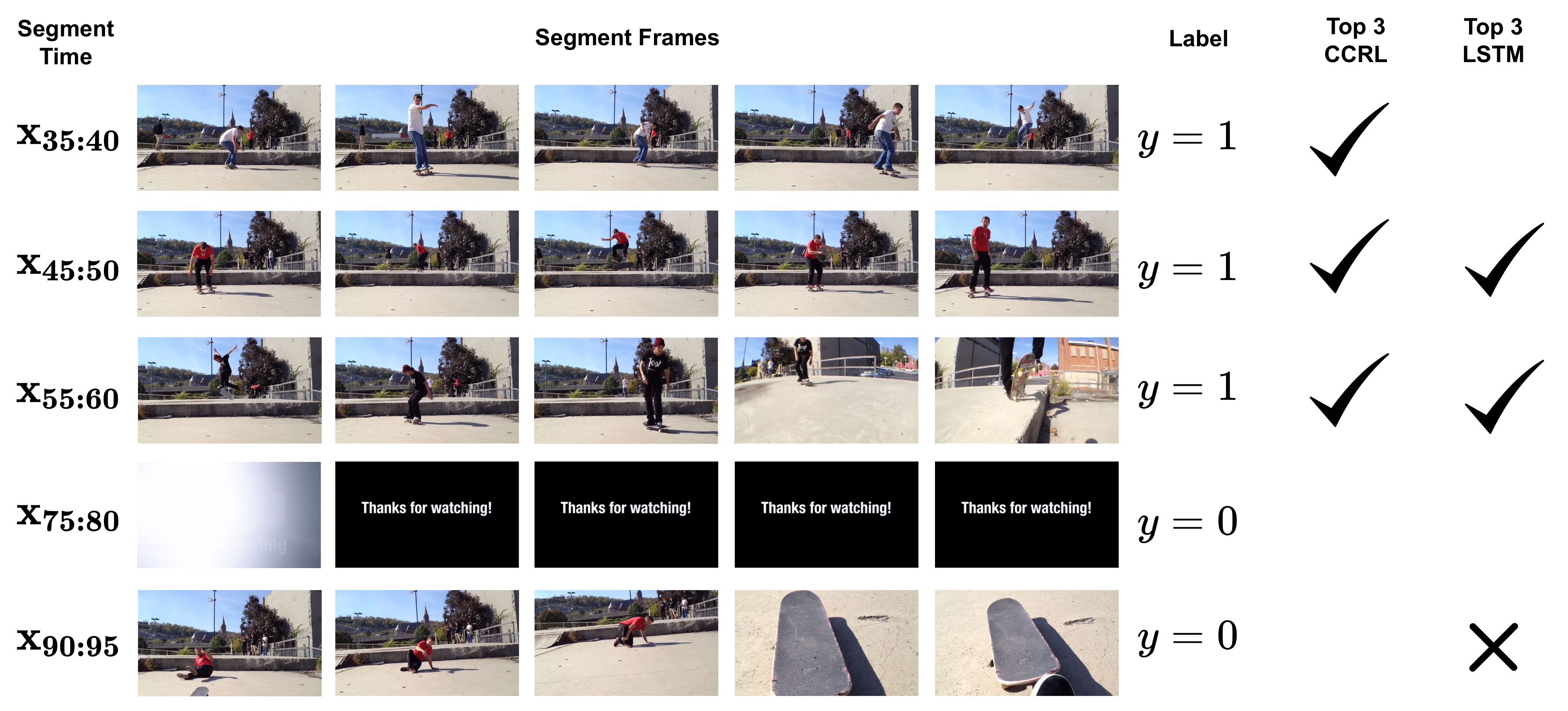}
\caption{Qualitative example of instance localization for \ccrl and LSTM models. 
Top-3 segments for class ``skateboarding'' are shown for each model. LSTM
makes a mistake on segment $\mathbf{x}_{90:95}$ where frames resemble skateboarding
but actually show a skateboarder falling. \ccrl is able to correctly identify all
relevant segments.}\label{fig:qualitative}
\vskip -0.5cm
\end{figure*}

For \ccrl, we use tree-based gradient boosting machines (GBMs)
to learn non-smooth class relationships that are difficult to
capture with deep learning. We use predictions from segment 
candidate generation models as additional input features for 
each segment. Class features include one-hot class encoding
and various versions of similarity features outlined in 
Section~\ref{sec:features}. We use the XGBoost~\cite{chen2016xgboost} 
package for GBM training in all experiments.

\subsection{Results}

Table~\ref{table:candidate_recall} summarizes results for segment candidate generation
models. We focus on recall here since the primary aim for candidate generation is
to capture as many relevant segments as possible. From the table we see that candidate generation is able to reduce total number of segments from around 2.2M to 100K
and still achieve near perfect recall. The model ensemble in particular is  
able to achieve recall of 0.9969 which indicates that virtually all relevant 
segments are included in the candidate set.

Table \ref{table:segment_rerank} compares leaderboard model performance for all segment models 
before and after applying candidate generation. We observe significant increase in performance with up to $20\%$ gain in mAP after candidate generation. This indicates that candidate generation is able to successfully transfer information from the much larger video classification dataset and improve localization. Notably, models that perform worse on their own, such as
transformer, benefit more from the candidate generation.
We also see that \ccrl is the best performing model both with and without
candidate generation. Strong performance suggests that \ccrl is a promising
approach for concept localization and potentially other related tasks.
Model ensemble further improves performance and gives our best accuracy of
0.8329. We experimented with various ensemble techniques but found 
that simple averaging worked best here.

Figure~\ref{fig:ap} shows average precision performance for \ccrl and LSTM models 
for a subset of classes clustered according to their bag-of-words descriptions. 
Here, we can see the benefits of cross-class learning. For example, LSTM performs
significantly worse for classes ``BMW 3 Series (E30)'' and ``Samsung Galaxy'',
while related classes have higher performance. However, \ccrl is able to achieve 
more consistent accuracy across classes in a given cluster. One explanation for the more consistent
scores can
be attributed to better information sharing between classes. However, 
also observe that for some clusters such as the Xbox products, 
our model struggles to achieve consistent performance. Further architectural 
improvements or additional features can be beneficial here and we leave it
for future work.

Table~\ref{table:ranks} shows the final leaderboard scores for top-5 teams
as well as individual models. Our team ``Layer6 AI" achieved first place
outperforming the second team by 0.67 points in mAP. We also see that individual
models have strong performance and all place in the top-10 on the leaderboard.
Our best single model places 5'th on the leaderboard and can be used on its
own, particularly in production environments where model ensembles can
be difficult to manage.

Figure~\ref{fig:qualitative} shows a qualitative example comparing
CCRL and LSTM predictions on one video. Each row represents one of
five labelled segments for this video with class ``skateboarding". 
To evaluate localization, we examine the top-3 predicted segments from
each model. From the fiigure we see that LSTM makes a mistake on segment 
$\mathbf{x}_{90:95}$ where frames resemble skateboarding but actually 
show a skateboarder falling. \ccrl is able to correctly identify all
relevant segments.

\section{Conclusion}

We propose cross-class relevance learning approach for temporal
concept localization. Our approach uses a pairwise model that takes 
both segment and target class as input and predicts relevance. This
promotes information sharing between classes, and allows for extensive 
class feature engineering. We apply our approach to the YouTube-8M video
understanding challenge together with other leading models and achieve first place.
In the future, we aim to explore additional features to describe classes
as well as pairwise architectures for joint prediction.

{\small
\bibliographystyle{template/ieee}
\bibliography{yt}
}

\end{document}